%% file: root.tex
\documentclass[letterpaper, 10 pt, conference]{ieeeconf}  

\IEEEoverridecommandlockouts                              

\usepackage[dvipsnames]{xcolor}

\newcommand{\new}[1]{\textcolor{black}{#1}}

\title{\LARGE \bf
\modelname: Hyperbolic Planning and Curiosity for Crowd Navigation
}

\usepackage{multirow}
\usepackage{soul}
\usepackage{amsmath}
\usepackage{amssymb}
\usepackage{graphicx}
\usepackage{xspace}
\usepackage{subfigure}
\usepackage{float}
\usepackage{booktabs,xcolor,colortbl}
\newcommand{\modelname}{Hyp$^2$Nav\xspace}
\newcommand{\HICM}{HyperCuriosity\xspace}
\newcommand{\planner}{HyperPlanner\xspace}
\definecolor{Gray}{gray}{0.9}
\usepackage{fontawesome}
\usepackage{hyperref}

\author{
Guido M. D'Amely di Melendugno*$^1$\ \ \ 
Alessandro Flaborea*$^{1,2}$\ \ \ 
Pascal Mettes$^3$\ \ \
Fabio Galasso$^1$
\thanks{$^{*}$Authors contributed equally.
        }%
\thanks{$^{1}$Sapienza University of Rome, Italy, email:
        lastname@di.uniroma1.it}%
\thanks{$^{2}$ItalAI (italailabs.com)}%
\thanks{$^{3}$University of Amsterdam, Netherlands, email:
        P.S.M.lastname@uva.nl}%
}

\begin{document}



\maketitle



\input{Figures/scripts/navigation}



\input{Sections/1-Abstract}
\input{Sections/2-Introduction}
\input{Sections/3-Related}
\input{Sections/Background}
\input{Sections/4-Method}
\input{Sections/5-Results}
\input{Sections/6-Ablation}
\input{Sections/8-Limitations}
\input{Sections/7-Conclusions}










{\small
\bibliographystyle{IEEEtran}
\bibliography{biblio}
}

\end{document}

%% file: Figures/scripts/navigation.tex
\makeatletter
\g@addto@macro\@maketitle{\begin{figure}[H]
\    \centering
    
    \subfigure{\includegraphics[trim={2.8cm 3cm 2.8cm 3cm}, clip, width=0.3\linewidth]{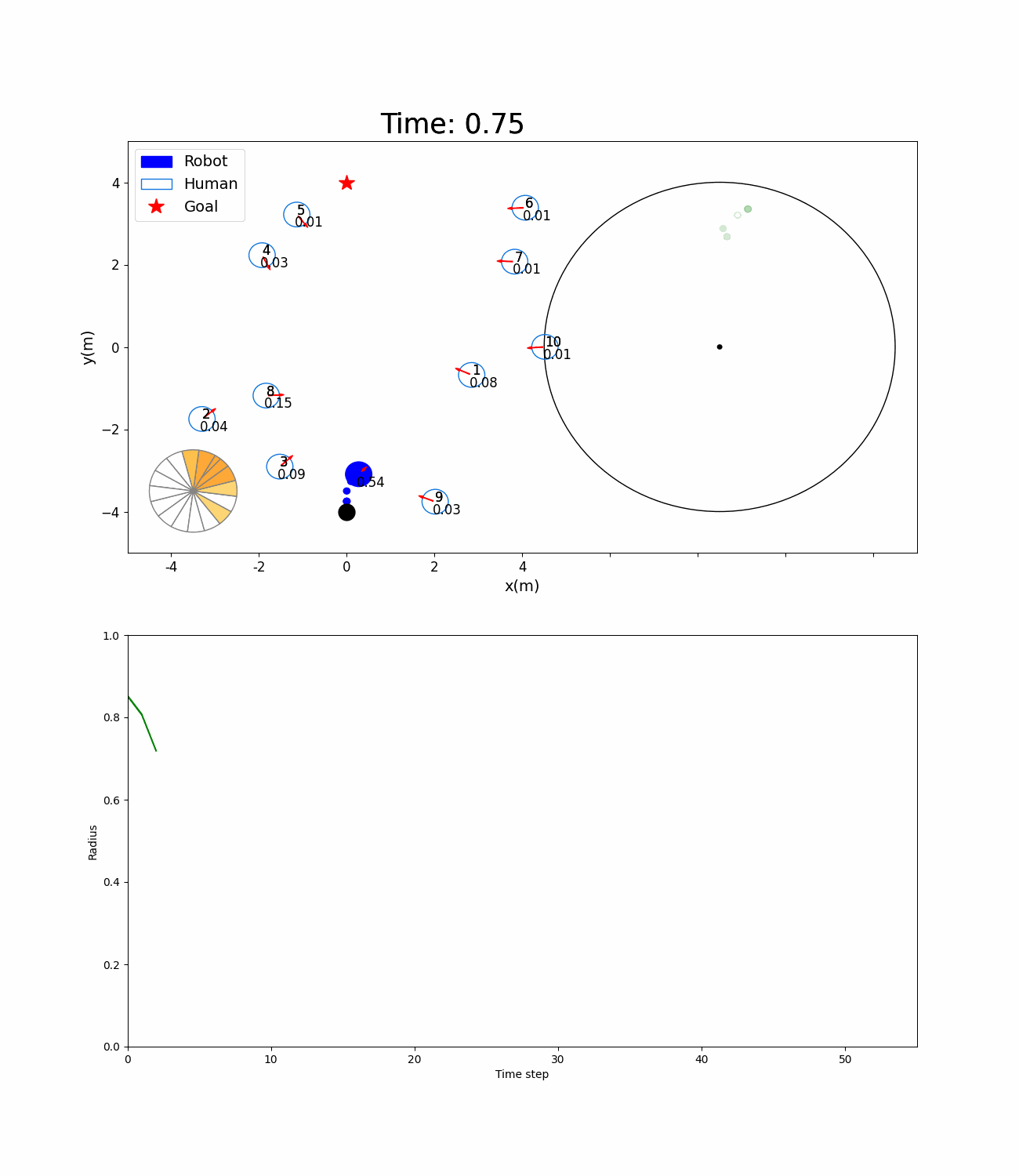}} 
    \subfigure{\includegraphics[trim={2.8cm 3cm 2.8cm 3cm}, clip, width=0.3\linewidth]{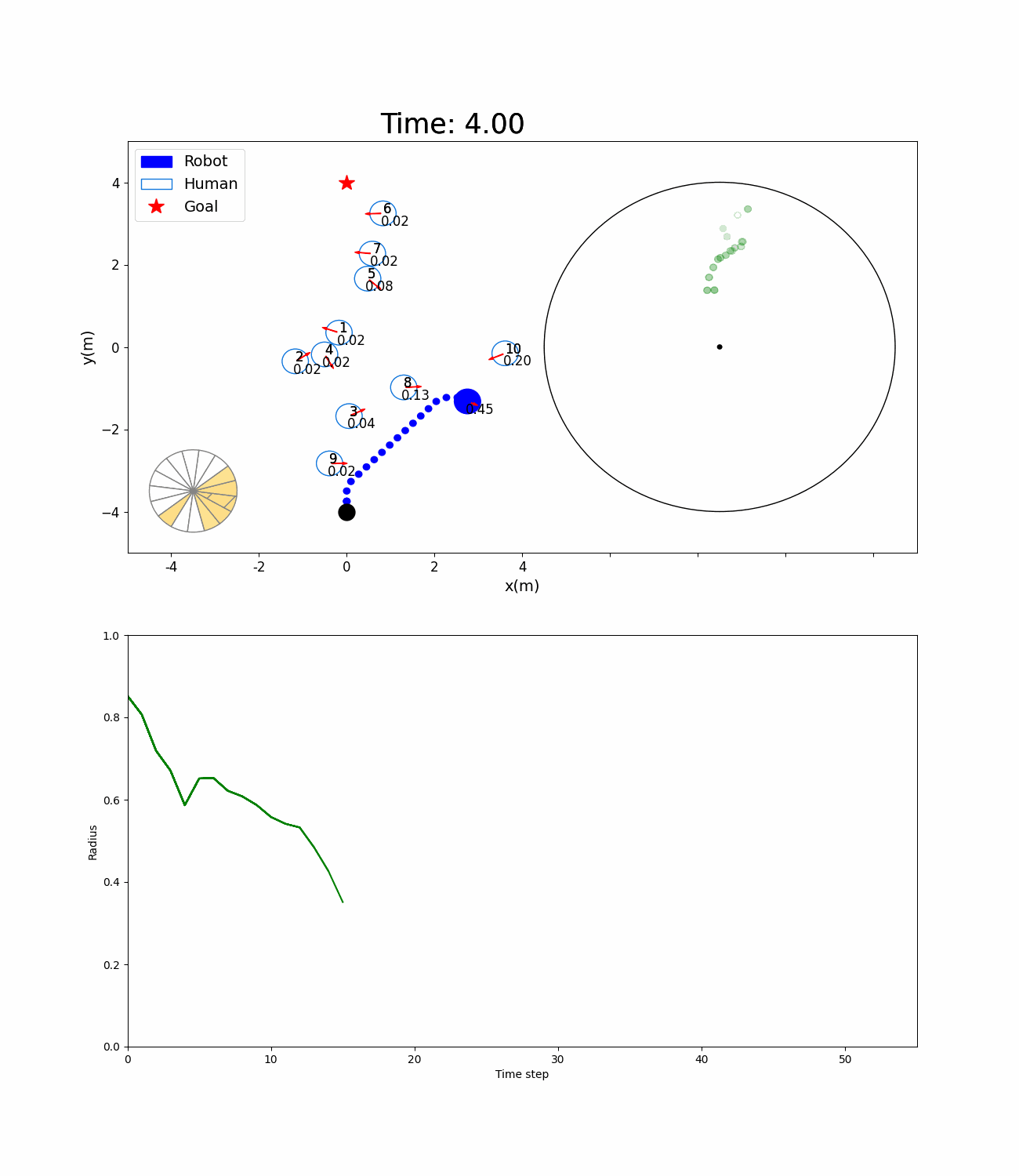}} 
    \subfigure{\includegraphics[trim={2.8cm 3cm 2.8cm 3cm}, clip, width=0.3\linewidth]{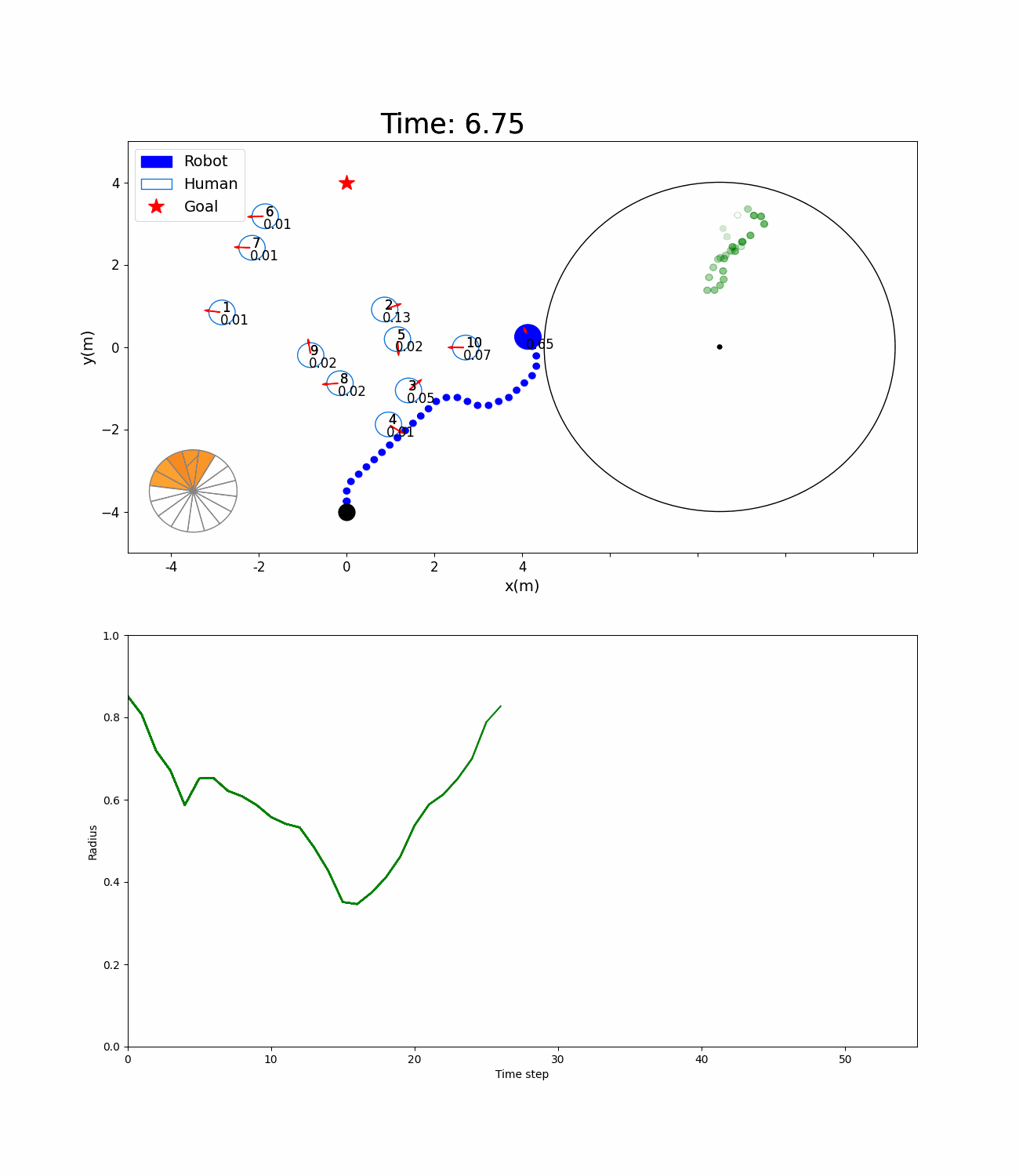}} 
    \caption{Navigation in the hyperbolic space}
\label{fig:visualization}
\end{figure}
}

%% file: Sections/1-Abstract.tex
\begin{abstract}
Autonomous robots are increasingly becoming a strong fixture in social environments.
Effective crowd navigation requires not only safe yet fast planning, but should also enable interpretability and computational efficiency for working in real-time on embedded devices.
In this work, we advocate for hyperbolic learning to enable crowd navigation and we introduce \modelname. Different from conventional reinforcement learning-based crowd navigation methods, \modelname leverages the intrinsic properties of hyperbolic geometry to better encode the hierarchical nature of decision-making processes in navigation tasks. 
We propose a hyperbolic policy model and a hyperbolic curiosity module that results in effective social navigation, best success rates, and returns across multiple simulation settings, using up to 6 times fewer parameters than competitor state-of-the-art models. 
With our approach, it becomes even possible to obtain policies that work in 2-dimensional embedding spaces, opening up new possibilities for low-resource crowd navigation and model interpretability. Insightfully, the internal hyperbolic representation of \modelname correlates with how much \textit{attention} the robot pays to the surrounding crowds, e.g.\ due to multiple people occluding its pathway or to a few of them showing colliding plans, rather than to its own planned route. \new{The code is available
at \href{https://github.com/GDam90/hyp2nav}{https://github.com/GDam90/hyp2nav}.}

\end{abstract}

%% file: Sections/2-Introduction.tex
\section{Introduction}\label{sec:intro}


Crowd navigation is paramount to deploying robots in environments shared with people~\cite{singamaneni2024survey}. Most recently, robots are finding applications in hospitals,
navigating busy corridors to transport medications, 
in robot postal services, restaurants and hotels. Across these applications, robots require advanced social navigation capabilities, which are yet to be acquired.
Thus,
the ubiquity of robots in settings traditionally occupied by humans has spurred a significant body of research~\cite{fiorini1998,van2008orca,chen2019crowdnav,helbing1995Social,xu2023} focused on enhancing their autonomy and interaction capabilities to ensure a safe human-robot co-existence. Among the compelling challenges stemming from this paradigm, the problem of robot navigation in crowded environments is critical to guarantee safe and seamless integration~\cite{zhou2022robot,zhou2023safe}.

For social navigation in crowds, the challenge is to guarantee a minimum travel time while ensuring maximum comfort/safety for human co-inhabitants.
This task is further complicated by computational constraints, the unpredictability of human behavior, and the need for an optimal representation of decision-making processes, which currently remain open problems. \input{Figures/scripts/teaser}
We observe that the efficiency requirement conflicts with the high-dimensional representations typically employed by (Euclidean) Deep Reinforcement Learning (DRL). Such high-dimensional representations are currently needed for effectively modeling states in crowd navigation but lead to excessive memory usage. 
Also, the complexity of human behavior may result in abrupt changes in the plans of the crowds and, eventually, collisions, which demands increased interpretability of the robot's decisions. 
Finally, a DRL path decision process resembles a hierarchical tree of the agent's internal statuses, and its representation tends to suffer from distortion in conventional Euclidean latent spaces~\cite{peng2021hyperbolic}. State-of-the-art works employ conventional neural networks, which are de facto operating in Euclidean space
\cite{zhou2023safe,martinez2023}. Here, we argue that the global state of the environment and the inherent Markov Decision Process (MDP) are intrinsically graphs, inconvenient for Euclidean spaces~\cite{ganea2018hyperbolic}.
Hyperbolic learning holds significant potential, which the state-of-the-art~\cite{zhou2023safe,martinez2023} does not leverage.

We propose a novel hyperbolic path-planner with hyperbolic curiosity within a deep reinforcement learning framework, which we dub \emph{Hyp$^2$Nav}. 
Thanks to hierarchically-organized decision processes, endowed by the hyperbolic model, \emph{Hyp$^2$Nav} increases success rates and rewards at a fraction of the parameter count.
Our approach is inspired by advances in
Hyperbolic Neural Networks (HNNs), which have recently garnered attention in computer vision, graph networks, recommender systems, and more for embedding hierarchical data due to their inherent property of embedding tree-like structures with minimal distortion. 
We advocate for adopting hyperbolic latent spaces to represent the states in an MDP,
proposing a hyperbolic policy module, \emph{\planner}, to operate entirely in low-dimensional latent hyperbolic spaces, for the first time exploiting extremely lightweight \emph{2-dimensional} representations.
We furthermore present \emph{\HICM}, to enforce exploration during training in hyperbolic space. Coupling the benefits of the recent \emph{Intrinsic Curiosity Module} (ICM)~\cite{pathak2017ICM} with the novel hyperbolic states representations, \HICM reports increased exploration, especially in the early training episodes, speeding up convergence to the optimal policy.
\modelname yields accurate and generalizable policies, as we demonstrate in comparative evaluation.

We also investigate the inner working of \modelname, identifying interesting properties of the hyperbolic radius, i.e., the magnitude of the hyperbolic embedding.
We show in our experiments that the radius of the representations correlates with the current navigation's complexity, the collision's danger, and, thus, the robot's uncertainty. 
Fig.~\ref{fig:teaser} supplies an example of this, showing that the hyperbolic radius decreases as the robot gets closer to humans (\textit{top-left} and \textit{bottom-right} frames) and increases when the path is clear of obstacles (\textit{top-central} and \textit{top-right} frames).

Following established literature~\cite{chen2019crowdnav,xu2023,zhou2022robot,martinez2023}, we assess our approach by benchmarking it against state-of-the-art techniques with fair simulations in complex and simple scenarios.
Overall our main contributions are as follows:
\begin{itemize}
    \item We propose \emph{Hyp$^2$Nav}, the first DRL hyperbolic path-planner with hyperbolic curiosity that features a hierarchical path decision process;
    \item \emph{Hyp$^2$Nav} maintains high success rates and returns with low-dimensional embeddings, as low as 2;
    \item We find novel interpretable properties of the radius embedding norm, said hyperbolic radius.
\end{itemize}

%% file: Figures/scripts/teaser.tex
\begin{figure}[t]
    \centering
    
    \includegraphics[trim={2.1cm 2.2cm 2.5cm 2.5cm}, clip, width=0.9\linewidth]{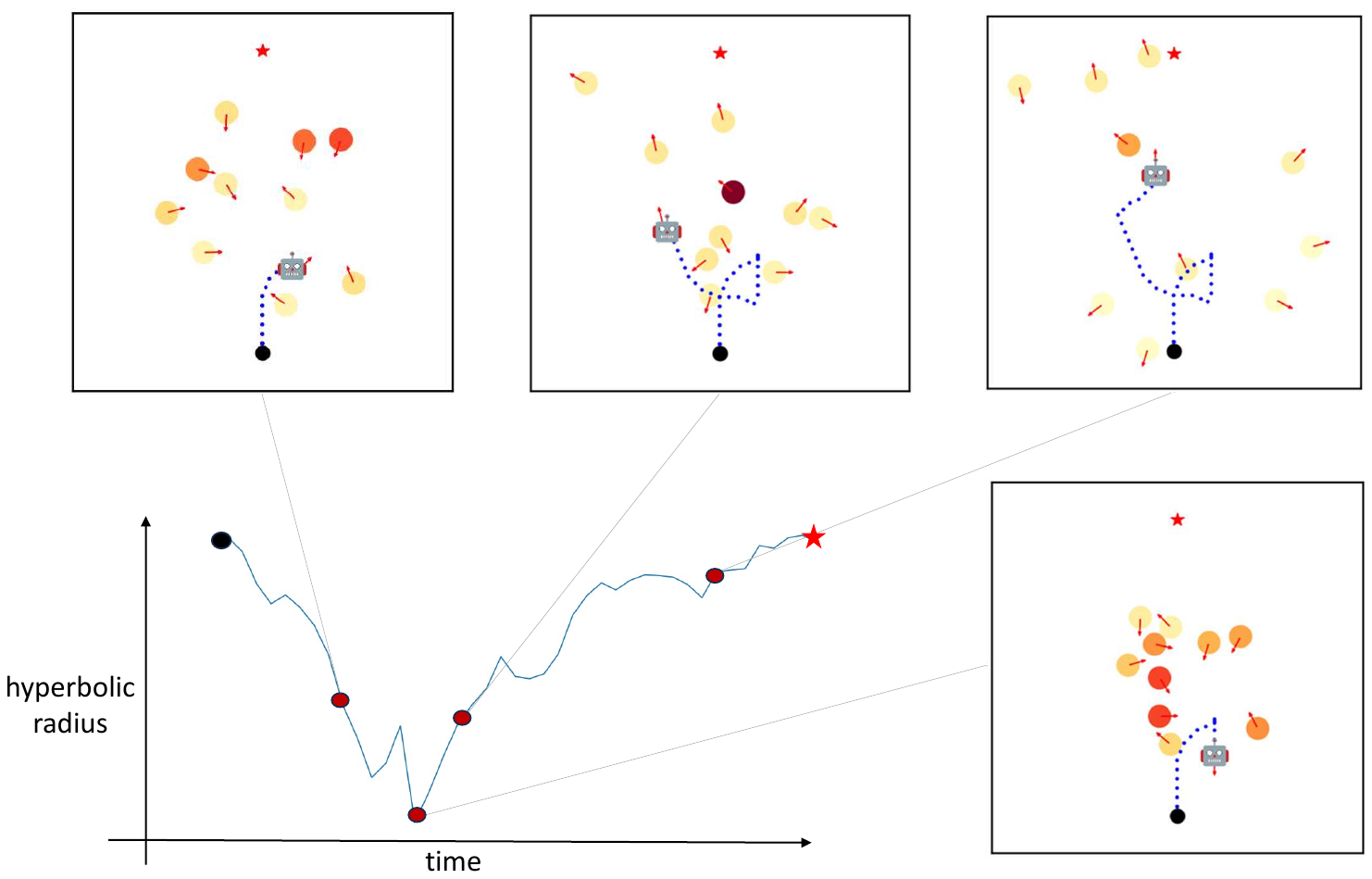}
    \caption{\textbf{Visualizing the hyperbolic radius}, the magnitude of the hyperbolic embeddings.
    Optimizing policies for crowd navigation in hyperbolic space results in embeddings that correlate with the robot's uncertainty in navigating within the obstacles in the scene.
    In the bottom left plot, the value of the hyperbolic radius (\textit{y-axis}) is depicted over time (\textit{x-axis}) in a typical rollout.
    The radius decreases when encountering obstacles directed towards the robot (\textit{top-left box}), indicating reduced confidence in the robot's decisions. Notice that people are color-coded circle, the \textit{red}-er, the larger is the robot attention to them (instead of self). Conversely, as the robot successfully navigates through challenging scenarios (\textit{bottom-right}), the hyperbolic radius increases (\textit{top center}), reflecting improved confidence and more straightforward decision-making toward the final goal \textcolor{red}{\faStar} (\textit{top right}).
    }
\vspace{-1.5mm}
\label{fig:teaser}
\end{figure}

%% file: Sections/3-Related.tex
\section{Related Works}
\label{sec:rw}


\subsection{Crowd Navigation}
\label{rl_socnav}
Early studies in crowd navigation~\cite{helbing1995Social,ferrer2013ExtendedSocial,van2011ORCA} propose models describing agents' behavior within crowds. These studies introduce fundamental concepts such as {\it Social Forces}~\cite{helbing1995Social} and provide models that guarantee collision avoidance in case of perfect information even for multiple agents simultaneously~\cite{van2011ORCA}.
The approaches only employ past information about the crowd motion and do not explicitly model the future trajectories of the obstacles.
A second line of work~\cite{alahi2016lstmrl,katyal2020,salzmann2020} focuses on predicting the trajectories of the dynamic obstacles to devise a safe path for the robot. While effective in terms of predictions, these techniques require a lot of computation, especially when dealing with large crowds, causing the decision-making model to be too slow for real-time applications. Chen et al.~\cite{chen2017CADRL} describe the decision-making problem as an MDP and 
proposes to adopt DRL 
to solve the task. Following this approach, Chen et al.~\cite{chen2019crowdnav} extend the agent's attention from human-robot interaction to a more comprehensive crowd-robot interaction, explicitly modeling the human-robot and human-human interactions. SG-D3QN~\cite{zhou2022robot} introduces a social attention mechanism to retrieve a graph representation of the crowd-robot state, which can be further improved~\cite{martinez2023,xu2023} by relying on intrinsic rewards~\cite{pathak2017ICM,seo2021RE3} and spatio-temporal maps for environment representations. Our work takes inspiration from Martinez et al.~\cite{martinez2023} and introduces hyperbolic latent representations to encode the MDP states, enabling us to achieve high performance with low-dimensional embeddings.

Exploration in Crowd Navigation is paramount for the autonomous agent to learn nuanced decision-making, balancing the need to navigate close to obstacles for efficiency with ensuring safety in populated environments.
ICM~\cite{pathak2017ICM} and RE3~\cite{seo2021RE3} are recent methods that implement this strategy providing the policy network with ``bonus rewards'' when the agent visits unknown spaces. 
In ICM, a small network is fed with the current state and the action the agent is performing, and it is tasked to predict the representation of the subsequent state. The more the error (meaning that the state is unknown), the more the intrinsic reward. This work extends ICM, adopting hyperbolic state representations to increase further the reward for exploring novel states. 

\subsection{Hyperbolic Neural Networks}
\label{rl_hyper}
Hyperbolic Neural Networks (HNNs) are emerging as a powerful tool for capturing hierarchical representations, yielding compact representations, dealing with uncertainty, and much more~\cite{mettes2023hyperbolic,peng2021hyperbolic}.
Recent works in computer vision showcase the potential of HNNs, outperforming traditional Euclidean models~\new{\cite{desai2023meru,vanspengler2023poincare}} in several tasks, \new{such as image segmentation~\cite{atigh2022,halo}, anomaly detection~\cite{hypad}, and pose forecasting~\cite{hysp}}. Recently, Cetin et al.~\cite{cetin2023hyperbolic}, 
acknowledge that MDPs can be represented as tree graphs in the states space, thus they advocate for using hyperbolic latent representations in the domain of DRL, as they natively extract hierarchical features from the data. They demonstrate the advantages of coupling hyperbolic learning with DRL methods on Procgen~\cite{cobbe2020procgen} and Atari-100k~\cite{bellemare2013atari}, reporting performance improvements on a wide range of test environments. They define a hybrid network with an Euclidean backbone and hyperbolic projective and classification layers. This work, for the first time
applies the insights of Cetin et al.~\cite{cetin2023hyperbolic} to the crowd navigation problem.
We furthermore introduce new value and curiosity modules operating entirely in hyperbolic space.
Moreover, we investigate the efficacy of hyperbolic intrinsic rewards and interpretability by analyzing hyperbolic features.

%% file: Sections/Background.tex
\section{Background}\label{sec:bg}

\subsection{Problem formulation}\label{method_pf}
The crowd navigation problem consists of finding the optimal policy that drives an autonomous agent to a goal position in a crowd.
For our simulation, we leverage the widely adopted CrowdNav simulator~\cite{chen2019crowdnav}, see e.g. ~\cite{chen2019crowdnav,zhou2022robot,xu2023,martinez2023,chen2017CADRL,everett2021LSTMRL}.
In this environment, the dynamic obstacles are informed of each other agent's state (apart from the robot's state) to avoid reciprocal collisions. Those states encompass their current position~($p\in \mathbb{R}^2$), their velocity~($v\in \mathbb{R}^2$), and the radius $r$ used as a 2D proxy of their volume. Thus, we describe the $i$-th obstacle visible state at time $t$ as:
\begin{equation}
    w_t^i=[p_x,p_y,v_x,v_y,r]
\end{equation}
and refer to the state of all obstacles as $w_t^h= [w_t^1, \cdots, w_t^n ]$
The robot is designed as a holonomic robot, and its state encompasses the current position~($p\in \mathbb{R}^2$) and velocity~($v\in \mathbb{R}^2$), its radius $r$, the maximum scalar velocity $v_M$, the current steering angle $\theta$ and the goal position~($g\in \mathbb{R}^2$), such that:
\begin{equation}
    w_t^r=[p^r_x,p^r_y,v_x,v_y,r,v_M,\theta]
\end{equation}
At each timestep, the robot senses the environment's visible state, i.e., $w_t=[w_t^r,w_t^h]$. The autonomous agent moves according to the policy provided by our RL algorithm, which allows the models to steer along 16 equally spaced angles in $[0,2\pi)$ and move at 5 different velocities or stay still, resulting in 81 possible actions in each state. In the following, we define $\mathcal{A}$ as the set of allowed actions for the robot.

\subsection{Reinforcement Learning for crowd navigation}\label{bg_rl}

Following leading approaches~\cite{chen2019crowdnav,xu2023,zhou2022robot,martinez2023}, we cast the problem of crowd navigation as an MDP, where at each timestep $t$ the robot has to perform the optimal action ($a_t^*\in\mathcal{A}$) according to the current visible state. The problem is set as finding the optimal deterministic policy~${\pi^*:w_t\mapsto a^*_t}$ that associates the optimal action to the state $w_t$ for every timestep $t$. For retrieving the optimal policy, we define $Q^*(w,a)$ as the optimal state-action function describing the expected value of taking a particular action $a$ being in a certain state $w$. $Q^*(w,a)$ satisfies the Bellman equation:
\begin{equation}\label{bellman}
    Q^*(w,a) = \sum_{w', r} \mathbf{P}(w', r|w,a) [r + \gamma^t\max_{a'}Q^*(w',a')]
\end{equation}
where $w'$ is the state following $w$ after performing the action $a$, $r$ is the extrinsic reward provided by the environment, and $\gamma\in (0,1)$ is the discount factor that adjusts the interset for future rewards. The optimal policy $\pi^*$ is then defined as:
\begin{equation}
    \pi^*(w_t) = \arg \max_a Q^*(w_t, a)
\end{equation}
From Eq.~\ref{bellman}, it follows that accurately defining the reward is a key factor for RL algorithms. The extrinsic reward ($r^e_t=r^e(w_t)$) should be shaped to encourage the agent to reach its goal fast while avoiding collisions and creating discomfort to the human dynamic obstacles. As done in~\cite{martinez2023}, we formalize it as follows:
\begin{equation}
r^e_t =
\begin{cases}
0.25 & \text{if goal reached} \\
-0.25 & \text{if collision} \\
-0.2\ d_t^g + \sum_{i=0}^{N} f(d^i_t) & \text{otherwise}
\end{cases}
\end{equation}
where $d_t^g$ and $d^i_t$ represent the current distance from the goal and from the $i$-th obstacle, respectively. The summation term promotes a safety constraint, increasing the reward if the robot maintains a minimal distance from each obstacle. Thus, we devise $f$ as:
\begin{equation}
f(d^i_t) =
\begin{cases}
d^i_t - 0.2 & d^i_t < 0.2 \\
0 & \text{otherwise}
\end{cases}
\end{equation}

\subsection{Hyperbolic Deep Learning}\label{method_bg}

This paper advocates for hyperbolic learning to perform crowd navigation. A hyperbolic metric space is a Riemannian manifold with constant negative curvature $-c$ (in this paper, we consider $c=1$ unless explicitly stated)~\cite{ganea2018hyperbolic}. Its definition encompasses several isometric models, among which we select the Poincaré ball $\mathbb{D}^n$, following recent literature~\cite{cetin2023hyperbolic,atigh2022,vanspengler2023poincare,hypad,ghadimi2021hyperbolic}.
The Poincaré ball is an open ball of radius $1$, $\mathbb{D}^n=\{x\in\mathbb{R}^{n+1}:\ ||x||<1\}$, endowed with the Riemannian metric:
\begin{equation}
    g_x^\mathbb{D}=\frac{2 I_n}{1-||x||^2}
\end{equation}
with $I^n$ an $n\times n$ Identity matrix. 
Important in this paper is the ability to map points between hyperbolic space and its tangent space (i.e. Euclidean space), which are given by the exponential and logarithmic mapping functions.
Given a point $v\in\mathbb{D}^n$, the exponential map with basepoint $v$ $\exp_v$ projects any point from the tangent space of $\mathbb{D}^n$ in $u$ $u\in T_v(\mathbb{D}^n)$ into $\mathbb{D}^n$:
 \begin{equation}\label{expmapv}
    \exp_v(u)=v\oplus (\tanh(\frac{||u||}{1-||v||^2})\frac{u}{||u||})
\end{equation}
where $\oplus$ represent the Möbius addition~\cite{ganea2018hyperbolic}. As a common practice, we consider the origin $O$ of the Poincaré ball to be the basepoint for projections. This simplifies Eq.~\ref{expmapv} to:
\begin{equation}\label{expmapO}
    	\exp_O(u)=\tanh(||u||)\frac{u}{||u||}
\end{equation}
Inversely, the logarithmic map with basepoint $O$, $\log_O$, is defined as:
\begin{equation}\label{logmapO}
	\log_O(v)
        =
        \tanh^{-1}(||v||) \frac{v}{||v||}
\end{equation}
The distance between points on the Poincaré ball is:
\begin{equation}\label{PoincareDist}
    d_\mathbb{D}(x,y) = \mathrm{arcosh}\ {(1+\frac{2||x-y||}{(1-||x||^2)(1-||y||^2)})}
\end{equation}
Eq.~\ref{PoincareDist}, highlights the growth of the volume in hyperbolic space: when a point approaches the boundary of $\mathbb{D}^n$, its norm grows exponentially.
Leveraging the base operations described in Ganea et al.~\cite{ganea2018hyperbolic}, namely Möbius addition and Möbius matrix multiplication, we can extend the main modules commonly used for deep learning. In this work, we adopt hyperbolic multi-layer perceptrons (h-MLPs), which rely on Möbius matrix multiplication and addition.

%% file: Sections/4-Method.tex
\section{Methodology}\label{sec:method}
\begin{figure}
    \centering
    \includegraphics[width=1\linewidth]{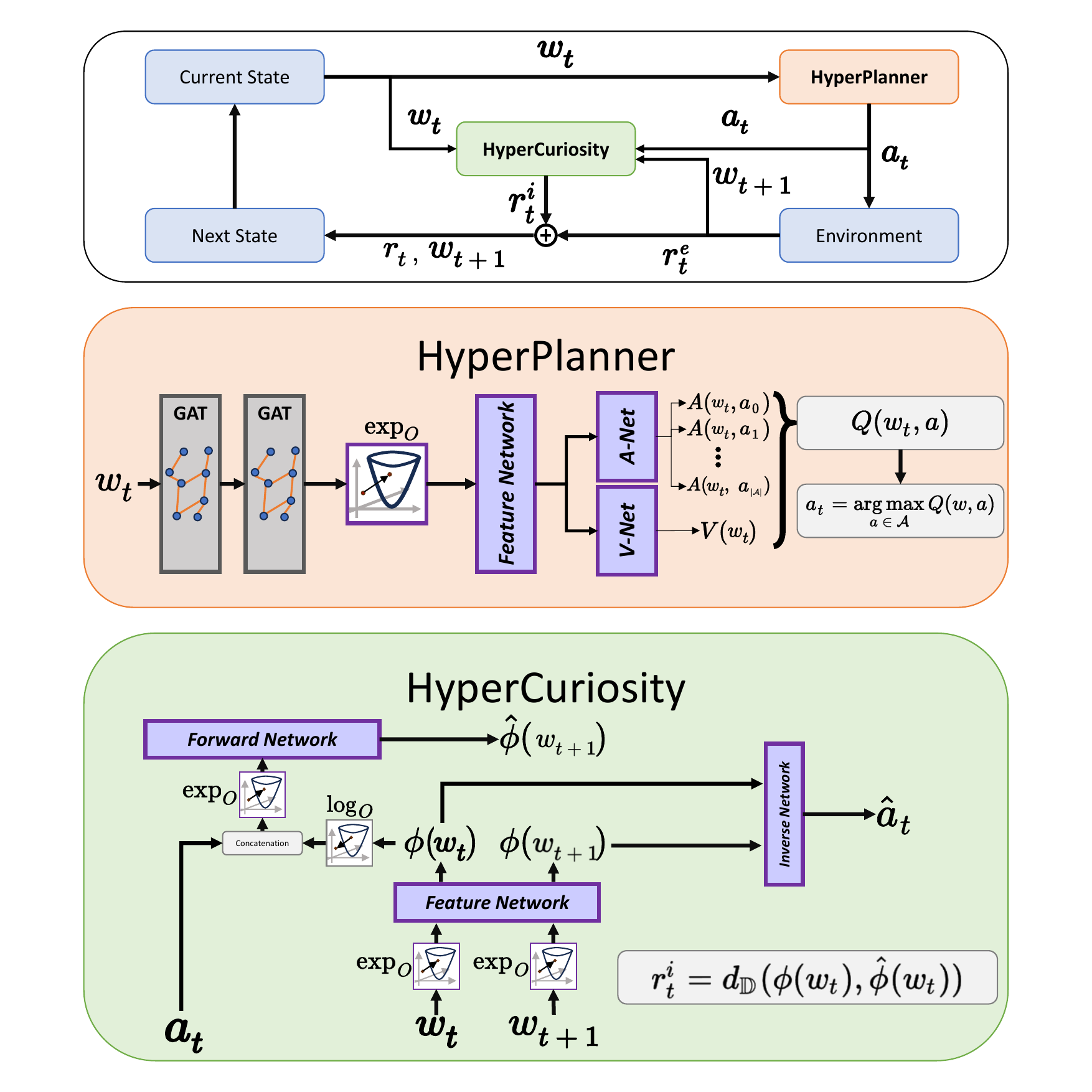}
    \caption{\textbf{Overview of \modelname.} We propose a hyperbolic policy network and a curiosity module to enable effective crowd navigation using only a few embedding dimensions. Modules in purple denote hyperbolic networks, \new{which are composed of several concatenated h-MLP layers}.}
    \label{fig:pictorial}
\end{figure}
In this work, we introduce \modelname. We propose a policy network, rooted in hyperbolic deep learning, dubbed \planner and responsible for taking optimal decisions (in Sec. IV-A), and a novel intrinsic reward module, dubbed \HICM and responsible for exploration (in Sec. IV-B). We outline both components 
below. An overview of our approach is shown in Figure~\ref{fig:pictorial}. 

\subsection{\planner}\label{method_sgdqn}
We introduce a model to navigate autonomous agents in dynamic environments. 
We employ graph encoding for the current state, followed by a value network adapted to fully operate in hyperbolic space that is tailored for graph representations and allows for extreme dimensionality reduction. Inspired by the Dueling-DQN~\cite{wang2016dueling}, the \planner focuses on 
determining the value of potential actions in each state, facilitating informed decision-making by the robot.

{\bf Environment graph.} We consider the environment's visible state $w_t$ as a graph whose nodes are represented by the agents in the scene, and the features of each node coincide with its observable state. 
This work relies on two subsequent graph attention (GAT) modules to encode this graph effectively. 
The first GAT layer receives as input an embedding $\Phi(w^*_t)$ of each agent's current state, obtained via an MLP $\Phi$ to have the same dimension for both $w_t^r$ ($\in\mathbb{R}^9$) and $w_t^i$ ($\in\mathbb{R}^5$), while the second layer outputs high-level representations along with softmax-normalized attention coefficient that describe the strength of the interaction of each agent with each other. 
We employ two GNN layers to account for second-order interactions, as the reciprocal interactions of two obstacles can affect the robot planning.

{\bf Hyperbolic Policy Network.} Next, we map the weighted representations to hyperbolic space via exponential mapping (cf. Eq.~\ref{expmapO}). A hyperbolic-MLP (h-MLP) module is in charge of extracting the compact hierarchical features and refining the coefficients given by the GATs. In the remainder of this section and in Fig.~\ref{fig:pictorial}, with a slight abuse of notation, we omit to represent the action of this module to keep the notations easy to read. 
Starting from these representations, two separate modules.
are tasked to (1) estimate the value of the current state (the \emph{V-Net}, $V$) and (2) quantify the relative advantage of each possible action on the current state (the \emph{A-Net}, $A$). 
The two modules are modeled as h-MLPs with ReLU activations:
\begin{equation}
\begin{split}
    V(w_t) &= \text{h-MLP}(\text{ReLU}(\text{h-MLP}(w_t))),\\ 
    A(w_t) &=[A(w_t, a^0),...,A(w_t, a^{|\mathcal{A}|})]\\ 
    &= \text{h-MLP}(\text{ReLU}(\text{h-MLP}(w_t)))
\end{split}
\end{equation}
Notably, aiming to reduce the computational overhead severely, we constraint these modules to operate in 2-dimensional latent spaces. Indeed, by leveraging the capacity of hyperbolic modules to encode data with a hierarchical structure, we rely on the compact yet expressive power of this reduced dimensional setting, allowing for efficient computation without sacrificing the depth and quality of the agent's environmental understanding.
Finally, we aggregate the outputs to retrieve a single value for each action in the current state as:
\begin{equation}
	Q(w,a) = V(w) + (A(w,a) - \frac{1}{\mathcal{A}} \sum_{a'\in\mathcal{A}} A(w,a') )
\end{equation}
and we choose the next action as
\begin{equation}
	a_t = \pi(w_t) = \arg\max_{a\in\mathcal{A}} Q(w,a)
\end{equation}

\subsection{\HICM}
\label{method_hicm}

Within our DRL framework, we propose a module, dubbed \HICM, that refines the exploration-exploitation trade-off by leveraging the distinctive metric properties of hyperbolic space. We build upon the concept of ICM~\cite{pathak2017ICM}, which, akin to human curiosity, encourages exploration by generating intrinsic rewards based on the prediction error of the consequences (next state) of the agent's action. Pathak et al.~\cite{pathak2017ICM} show that the intrinsic reward mechanism benefits from state representation invariants to factors that do not affect the agent's decisions.
\HICM further extends this study, adopting hyperbolic latent space to compactly encode the hierarchical features of the environment, \new{and leverages the inherent geometry of hyperbolic space to increase the reward for visiting novel states, encouraging the model to explore more during training}. 


\HICM consists of three interconnected components settled in hyperbolic space: a feature extractor $\phi$, a forward module $f$, and an inverse module $g$; its input corresponds to the triplet $(w_t, a_t, w_{t+1})$ representing the interaction between the agent and the environment produced by the current policy $a_t=\pi(w_t)$.
The feature extractor $\phi$ consists of an exponential mapping to embed the states into hyperbolic space and an h-MLP layer:
\begin{equation}
    \phi(w) = \text{h-MLP}(\exp_O(w))
\end{equation}
We apply $\phi$ to two subsequent states ($w_t\xrightarrow{a_t}w_{t+1}$) recovering their high-level representations $\phi(w_t), \phi(w_{t+1})$. Next, the forward module $f$, devised as an h-MLP, uses the current state's representation and the transition action $a_t$ (encoded as a one-hot vector) to predict the next state's representation provided by $\phi$. Since the concatenation involves two terms defined in different spaces, we first apply the $\log_O$ map (Eq.~\ref{logmapO}) to $\phi(w_t)$ to project it in the Euclidean space. We formalize this layer as:
\begin{equation}
\begin{split}
	f(w_t,a_t) &= \text{h-MLP}(\exp_O([\log_O{\phi(w_t)},a_t])) \\
	&= \hat{\phi}(w_{t+1})
\end{split}
\end{equation}
where $[\cdot,\cdot]$ is the concatenation operation.
The third component, $g$, acts as a regularizing module, using the representations from the feature extractor to infer the action that transitions between the two, indirectly modeling the information conveyed by $\phi$'s representations.
Besides using hyperbolic latent spaces to compactly represent the relevant features of the environment, our proposal lies in the intrinsic reward computation, which is derived from the discrepancy between $\phi(w_{t+1})$ and the predictive module's anticipated next-state representation $\hat{\phi}(w_{t+1})$: 
\begin{equation}
	r^i_t = d_\mathbb{D}(\phi(w_{t+1}), \hat{\phi}(w_{t+1}))
\end{equation}


{\bf Fully-Hyperbolic Architecture.}
The entire model architecture is hyperbolic, and it inherits its metric properties. 
The advantages of this choice are twofold. First, when representing hidden states of an MDP, the hyperbolic latent space is more appropriate as this space is naturally suited to embed tree-like structures, adhering to the hierarchical nature of consecutive decisions and state space decision-depending evolution~\cite{cetin2023hyperbolic}.
Second, 
the Poincaré distance sets a penalty that grows exponentially with the hyperbolic radius. This is reflected in an increased contribution of the intrinsic reward, effectively motivating the agent to explore novel states during training. Moreover, this property yields, as shown in Sec.~\ref{disc_hradius}, the opportunity for the model to increase exponentially the penalty for mistaken decisions when the situation is \emph{simple} and to reduce the penalty when the situation is \emph{complex}. Interestingly, when guided so, the hyperbolic radius of the embeddings is unsupervisedly learned, and its magnitude correlates with the situation's complexity.

All the components of \HICM are randomly initialized and trained end-to-end with the policy network. This scheme entails that the intrinsic rewards are more influential in the first part of the training and smoothly decrease throughout the training due to the optimization of the predictive network.
This represents an opportunity for the model to tune the exploration importance directly from data, which represents an intriguing alternative to the systematic decrease of exploration adopted in classical exploration strategies~\cite{martinez2023}.

\input{Figures/scripts/train_2}

{\bf Implementation details.} We train \modelname for 10k episodes \new{using two GTX 1080 Ti GPUs (the training takes $\sim$20 hours)} and then select the best checkpoint. We adopt RiemmanianAdam as the optimizer with a learning rate of ${10^{-3}}$. Fig.~\ref{fig:train_together} shows the training curves.


%% file: Figures/scripts/train_2.tex
\begin{figure}[t]
    \centering
    \subfigure{\includegraphics[width=0.48\linewidth]{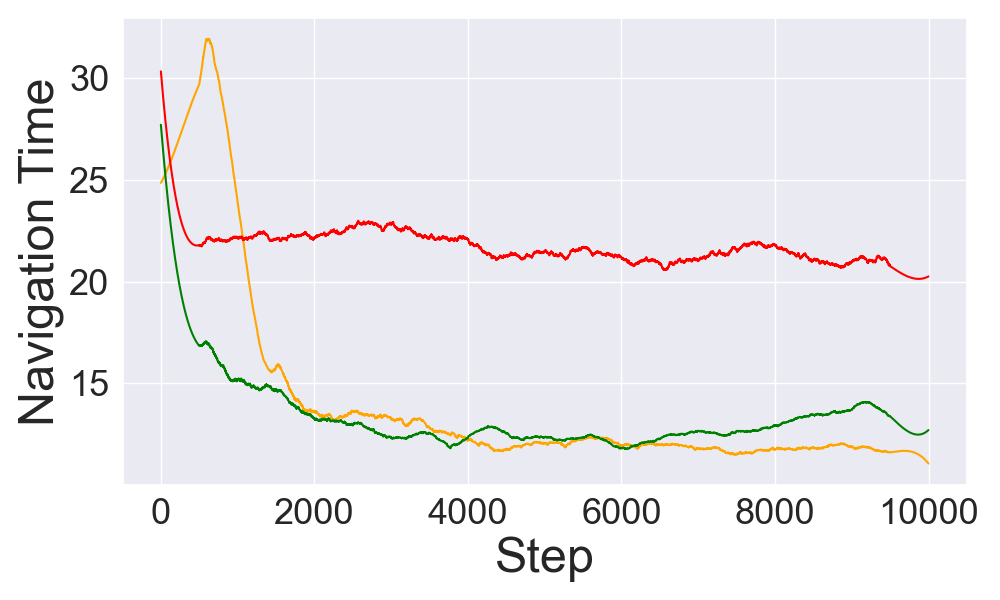}} 
    \subfigure{\includegraphics[width=0.48\linewidth]{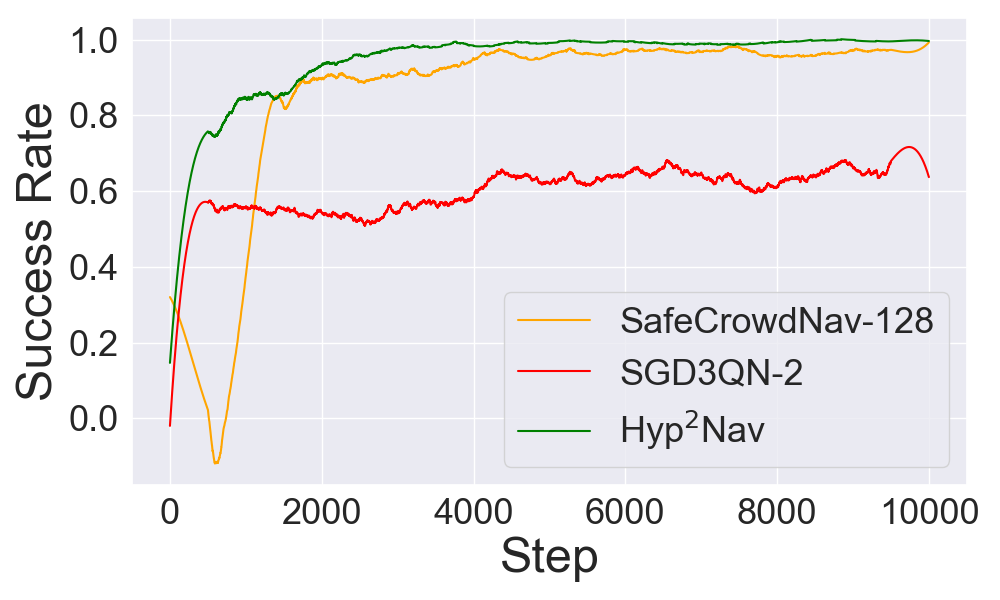}} 

    \caption{
    Learning curves during training of the navigation time and success rate of \modelname, SGDQ3N~\cite{martinez2023}, and SafeCrowdNav~\cite{xu2023}. The smoothing factor is 0.99.  
    }
\label{fig:train_together}
\end{figure}





%% file: Sections/5-Results.tex
\section{Experiments}\label{sec:experiments}

In this section, we first describe the benchmark we adopt, detailing the simulation environment, the baseline models, and the metrics (Sec.~\ref{exp_bench}).
Next, we discuss the results of \modelname in two increasingly complex scenarios (Sec.~\ref{exp_results}). 

\subsection{Simulation setup and benchmarking}\label{exp_bench}



{\bf Simulations.} To evaluate the proposed \modelname, we follow Zhou et al.~\cite{zhou2022robot} and use two scenarios from the CrowdNav~\cite{chen2019crowdnav} simulation environment: simple and complex. The simple scenario consists of 5 humans in the scene, positioned in a circle, who must reach a predefined goal by crossing the circle’s center. The complex scenario involves 10 humans, each with a predetermined objective, where 5 are positioned in a circle or a square, and the others are randomly set. The agent obtains a new random destination goal
upon reaching its current one. The ORCA~\cite{van2011ORCA} algorithm determines the human agents’ paths, allowing them to navigate without collisions. 
Following literature, we define the interval between two consecutive timesteps as 0.25 seconds. We train \modelname for 10000 episodes, evaluating its performance every 500 episodes.
Finally, we consider three conditions for ending an episode, namely collision(the robot collides with an obstacle), out-of-time (the robot fails to reach its destination within 30 seconds, but no collision occurs), and success (the robot safely reaches the goal within 30 seconds). 

\input{Tables/comparison_with_sota}

{\bf Baselines.} 
We benchmark our proposed \modelname against several state-of-the-art methods, which we adopt as comparative baselines. 
We consider ORCA~\cite{van2011ORCA}, which employs a reactive policy, setting pedestrian radii to $d_s = 0.2$ for preserving safety distances. Additionally, we evaluate against two versions of Intrinsic-SGD3QN~\cite{martinez2023}, ICM and RE3 \new{(reporting their results from~\cite{martinez2023} for the complex setting)}, which adapts the exploration modules from Pathak et al.~\cite{pathak2017ICM} to social navigation tasks. Furthermore, we assess SafeCrowdNav~\cite{xu2023}, the former state-of-the-art method, which utilizes intrinsic exploration rewards as in Martinez et al.~\cite{martinez2023} \new{and AEMCARL~\cite{wang2022adaptive} (only for the simple scenario)} and provides quantitative safety score assessments. We report the results of our solution for both the low (2) and high-dimensional (128) embedding versions.

{\bf Metrics.} 
We test each method on 1000 randomly generated episodes and evaluate them using three metrics: “Average Return”, the average cumulative return over steps, “Navigation Time”, the average time required for the robot to reach the goal, and “Success Rate”, the rate of the robot reaching the goal without collisions.

\subsection{Results}\label{exp_results}
We analyze the quantitative performance of \modelname with 
state-of-the-art approaches
on complex and simple scenarios, reporting the results in Tables~\ref{tab:comparison_10h} and~\ref{tab:comparison_5h}, respectively. 

\textbf{Complex Scenario.} Table~\ref{tab:comparison_10h} presents the evaluation on the complex scenario where the robot navigates through 10 humans to reach the final goal. Our proposal demonstrates higher success rates than other methods across the 2- and 128-dimensional versions. Notably, the 2-dimensional variant achieves a success rate of 99.3\%, surpassing the 128-dimensional version by 1.4\% and outperforming the best baseline by 2.9\% in the circle scenario, and by 1.2\% and 2.1\% in the square one. Moreover, our method achieves higher average returns than state-of-the-art methods, particularly with the low-dimensional version in both cases. Interestingly, although the 128-dimensional version reaches its destination $\sim$1 second faster on average than the 2-dimensional counterpart in both scenarios, \modelname consistently yields the best success rate, provoking fewer collisions.
The advantage of navigation in hyperbolic space is that this performance comes at a fraction of the parameters. Compared to the current state-of-the-art, our approach is 2.7$\times$ to 6.5$\times$ more parameter-efficient. 

\input{Tables/comparison_with_sota_5_new}

\textbf{Simple Scenario.}
Table~\ref{tab:comparison_5h} shows the evaluation results for the simple scenario. Similar to the complex case, our method demonstrates superior success rates and average returns for the 2- and 128-dimensional variants, showing improvements of 1.4\% and 5.6\%, respectively. Our solution achieves a perfect success rate in all 1000 test cases, reporting no collisions. Interestingly, in contrast to the complex scenario, the high-dimensional version outperforms the low-dimensional one in this case.
In the simple scenario, the Euclidean models appear more aggressive, as confirmed by the lower navigation time of~\cite{zhou2023safe,martinez2023}.
However, this also results in decreased safety, as the Euclidean-based models report substantial drops in success rate, more likely to expose potential humans to collisions.

%% file: Tables/comparison_with_sota.tex
\begin{table}[t]

\centering{

\caption{
\textbf{Comparison on the complex setting of CrowdNav~\cite{chen2019crowdnav}.} \modelname obtains the best performance at a fraction of the required number of learnable parameters.
}
\label{tab:comparison_10h}
\resizebox{\linewidth}{!}{
\begin{tabular}{lcccc} 
\toprule

 & \multicolumn{1}{c}{\scriptsize\textbf{Params.}}  & \multicolumn{1}{c}{\scriptsize\textbf{Nav. Time}$\downarrow$}  & \multicolumn{1}{c}{\scriptsize\textbf{Avg. Return}$\uparrow$} & \multicolumn{1}{c}{\scriptsize\textbf{Succ. Rate}$\uparrow$}  \\ 
\midrule

\textbf{Circle} & & &\\
\midrule
ORCA \cite{van2011ORCA} & -- &  11.01 & 0.331 & 76.9   \\
SGD3QN-ICM \cite{martinez2023} & 361K &  11.37  & 0.668 & 96.8  \\
SGD3QN-RE3 \cite{martinez2023} & 149K &  11.01 & 0.682  & 97.1  \\
SafeCrowdNav \cite{xu2023} & 361K & 11.18 & 0.673  & 97.7    \\
\rowcolor{Gray}
\modelname-128 & 144K & \bf{10.94} & 0.678  & 97.9    \\
\rowcolor{Gray}
\modelname & \textbf{55K} & 12.04  & \textbf{0.698} & \bf{99.3}   \\

\midrule
\textbf{Square} & & &\\
\midrule
ORCA \cite{van2011ORCA}  & -- & 12.86 & 0.442 & 84.0 \\
SGD3QN-ICM \cite{martinez2023} & 361K & 10.73 & 0.687  & 97.0 \\
SGD3QN-RE3 \cite{martinez2023}  & 149K & 10.22 & 0.675 & 95.8 \\
SafeCrowdNav \cite{xu2023} & 361K &  10.54 & 0.678   &  97.7   \\
\rowcolor{Gray}
\modelname-128 & 144K & \textbf{10.15} & 0.693   & 98.6     \\
\rowcolor{Gray}
\modelname & \textbf{55K} & 11.08 &  \textbf{0.715}  & \textbf{99.8}    \\

\bottomrule
\end{tabular}
}}
\vspace{-0.3cm}
\end{table}

%% file: Tables/comparison_with_sota_5_new.tex



\begin{table}[!t]
\centering{
\caption{\textbf{Comparison on the simple setting of CrowdNav~\cite{chen2019crowdnav}.} Akin to the complex setting, we are able to obtain the highest return and success rate with higher parameter efficiency.}
\label{tab:comparison_5h}
\resizebox{0.9\linewidth}{!}{
\begin{tabular}{lccc} 
\toprule
{\scriptsize\textbf{Method}} & \multicolumn{1}{c}{\scriptsize \textbf{Nav. Time}$\downarrow$}  & \multicolumn{1}{c}{\scriptsize \textbf{Avg. Return}$\uparrow$}  & \multicolumn{1}{c}{\scriptsize \textbf{Succ. Rate}$\uparrow$}  \\ 
\midrule
ORCA \cite{van2011ORCA}  &  13.87  & 0.323 & 73.6  \\
SGD3QN-ICM \cite{martinez2023} &  \textbf{9.79}   & 0.696 & 96.6\\
AEMCARL \cite{wang2022adaptive} &  12.86   & 0.539  & 92.0  \\
SafeCrowdNav \cite{xu2023} &  9.98   & 0.707 & 98.6   \\
\rowcolor{Gray}
\modelname-128 &  10.56   & \textbf{0.747} & \textbf{100} \\
\rowcolor{Gray}
\modelname  & 10.66    & 0.707 & 99.5  \\
\bottomrule
\end{tabular}
}
}
\vspace{-0.3cm}
\end{table}

%% file: Sections/6-Ablation.tex
\section{Discussion}\label{exp_discussion}
\input{Figures/scripts/barplots}

\input{Tables/main_results_20H}
\input{Figures/scripts/correlation}

\subsection{Embedding Dimensionalities}\label{disc_embdim}

Fig.~\ref{fig:barplots} illustrates the performance comparison between two methods, {\it Intrinsic-SGD3QN}~\cite{martinez2023} and our proposed approach \textit{\modelname}, across different embedding dimensions (2, 8, and 128).
In all the considered dimensions, our hyperbolic learner consistently outperforms its Euclidean counterpart, illustrating two critical advantages: the effectiveness of hyperbolic modules even with minimal embedding sizes and the overall convenience of adopting a hyperbolic approach for state representation in the crowd navigation task. 
Indeed, the Euclidean competitor fails to converge when employing only 2 dimensions, reporting a clear drop in both the average navigation time and success rate metrics.


In Fig.~\ref{fig:train_together}, we compare the training trends of \modelname, \new{SafeCrowdNav~\cite{xu2023},} and Intrinsic-SGD3QN~\cite{martinez2023}. The plot clearly shows that the Euclidean model\new{s underperform \modelname, and SGD3QN} fails to converge when the latent space is constrained to 
2 dimensions (\new{red} curve), reporting 
low success rate, and high average time, indicating that most episodes conclude matching an out-of-time condition.



\subsection{Hyperbolic Radius and Uncertainty}\label{disc_hradius}
While training its policy, \modelname learns to encode the different states composing an episode in different areas of the Poincaré Disk. 
We reach this conclusion upon investigation of \modelname's representations by analyzing their hyperbolic radius, which is the distance from the center of the Poincaré Disk to the embeddings. In Fig.~\ref{fig:correlation}, we plot the hyperbolic radius of the current state Vs the attention the system is paying to obstacles other than self for a single timestep. The plot reveals a high negative correlation between these measures (0.55). This confirms that more complex situations correspond to a smaller radius, indicating that the agent is more attentive to possible collisions. Consequently, the smaller radius reflects increased uncertainty.

On the sides of Fig.~\ref{fig:correlation}, we associate to some points the visualization of the current dynamic running in the simulation. In particular, we select two points with low radius ($\leq0.5$, red and purple points) and two with high radius ($\geq0.5$, yellow and green points). As can be seen from the Figure, we found that the points with lower radius values correspond to situations in which the obstacles are likely to impede the robot from easily moving forward. 
In the bottom-left frame, the robot cannot move toward the goal as a person is approaching on its left and cannot devise a safe root moving to the right, given the presence of a second obstacle on its right, demonstrating awareness of the situation's complexity.
Similarly, in the purple frame, the robot is surrounded by several obstacles moving toward the line between the agent and its goal. Still, the robot successfully avoids collision, but the safer way it has to take is opposite to the direction of the goal. On the other hand, the other two examples show the robot being sure of the next action, as no obstacle seems to approach the path to the goal. Surprisingly, the robot is near an obstacle in the bottom-right frame, but it correctly predicts its intention to move away. In the top-right one, the robot is mainly focused on obstacles that are far away from it. Still, they point to the optimal direction for the robot and risk colliding in a future step.

\subsection{Generalization}\label{disc_generalize}
We perform a study on the generalization capabilities of our model to further investigate the quality of \modelname's internal representations. For this experiment, we increase the scenario's complexity; Table~\ref{tab:together_20H} reports the results of the best performer models from Table~\ref{tab:comparison_10h} when doubling the number of dynamic obstacles in the scene. In particular, all the baselines represented in Table~\ref{tab:together_20H} have been trained in the complex scenario (10 obstacles) but are assessed with 20 humans. Even in this challenging scenario, \modelname shows the best performance for success rate and average return, outperforming the most recent model from~\cite{xu2023} by 4 p.p. points and surpassing the Intrinsic-SGD3QN (128-embedding dimensions). Although there is a slight increase ($\leq 1$ sec.) in navigation time compared to the best performer baseline, the trade-off is balanced by the gains in navigational safety and efficiency. These findings highlight the robustness and adaptability of \modelname, which is key in real-world crowd navigation when the number of entities composing the crowd is unknown and can vary with time. 

%% file: Figures/scripts/barplots.tex
\begin{figure}[t]
    \centering
    \subfigure{\includegraphics[trim={4cm 1.5cm 4cm 1.5cm}, clip, width=0.48\linewidth]{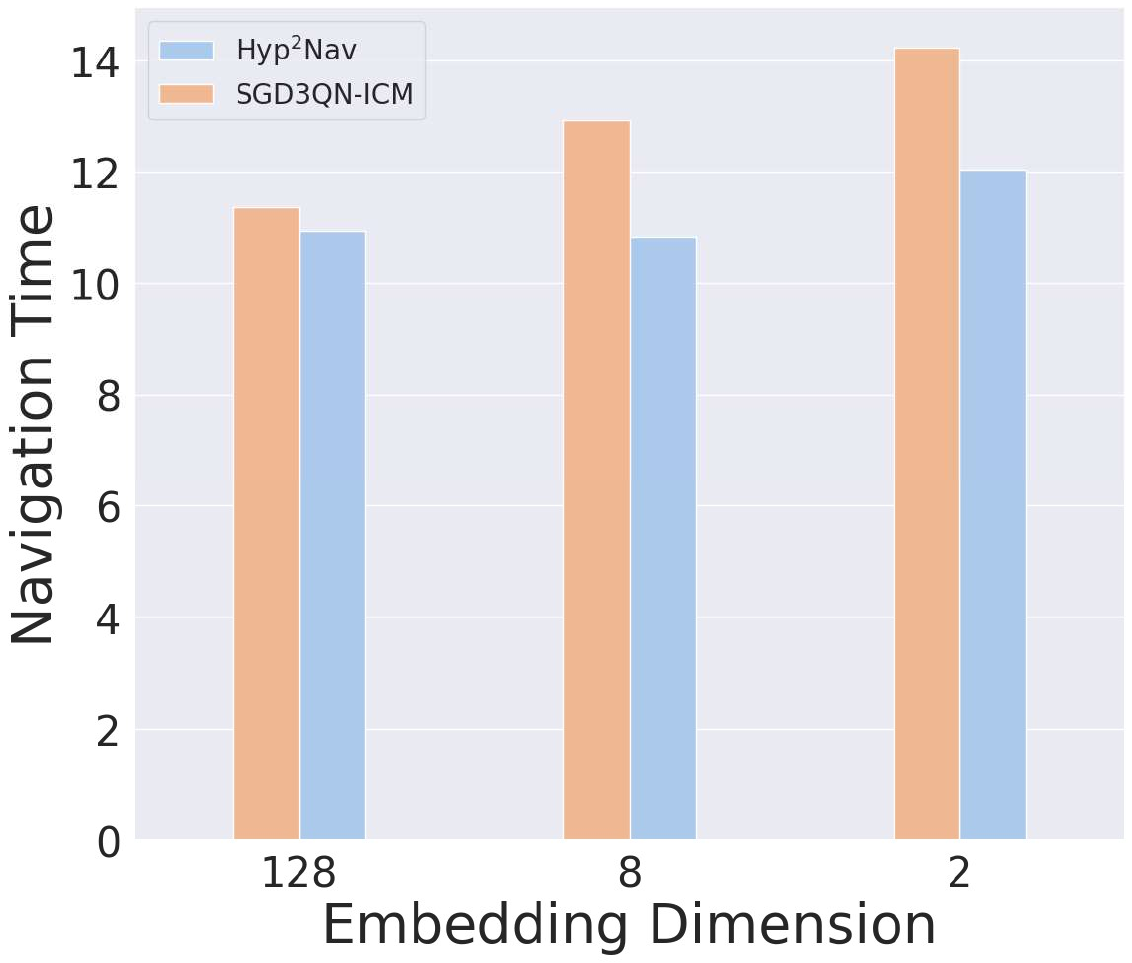}} 
    \subfigure{\includegraphics[trim={4cm 1.5cm 4cm 1.5cm}, clip, width=0.48\linewidth]{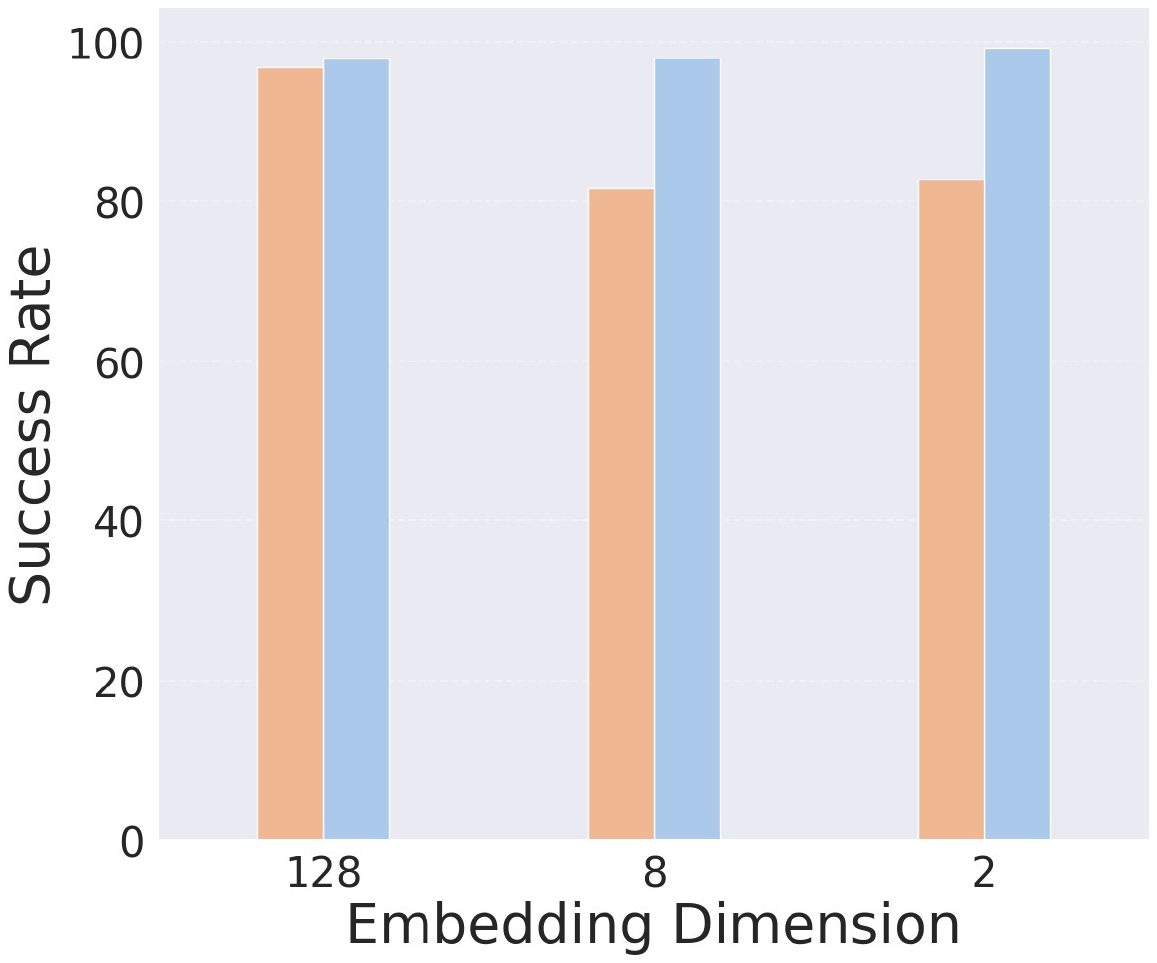}} 

    \caption{Comparison of the proposed \modelname and SGDQ3N~\cite{martinez2023} with different embedding dimensions. For both navigation time (left) and success rate (right), \modelname achieves better performance across all the dimensionalities. In the case of 2 and 8, SGDQ3N~\cite{martinez2023} fails to converge, whereas our proposed model shows the best performance with 2 dimensions. 
    }
\label{fig:barplots}
\end{figure}

%% file: Tables/main_results_20H.tex
\begin{table}[t]

\centering{

\caption{
\textbf{Comparison on a more complex setting of CrowdNav~\cite{chen2019crowdnav}.} Trained on the complex setting, \modelname achieves the best success rate and average return when doubling the obstacles in the scene.
}
\label{tab:together_20H}
\resizebox{\linewidth}{!}{
\begin{tabular}{lccc} 
\toprule

\textbf{Method} & \multicolumn{1}{c}{\textbf{Nav. Time} $\downarrow$}  & \multicolumn{1}{c}{\textbf{Avg. Return} $\uparrow$}  & \multicolumn{1}{c}{\textbf{Success Rate} $\uparrow$}   \\ 
\midrule

\multirow{1}{*}{SGD3QN\new{-ICM}~\cite{martinez2023}}  &  \textbf{13.81} & 0.559 &  93.2  \\
SafeCrowdNav\cite{xu2023} &  13.86  & 0.478 &   90.6  \\
\rowcolor{Gray}
\multirow{1}{*}{\modelname} &  14.55 & \textbf{0.571}  & \textbf{94.6}   \\

\bottomrule
\end{tabular}
}}
\vspace{-0.3cm}
\end{table}

%% file: Figures/scripts/correlation.tex
    

\begin{figure*}[h]
    \centering
    \vspace{0.5cm}

    \includegraphics[trim={0cm 4cm 0cm 5.5cm}, width=0.8\linewidth]{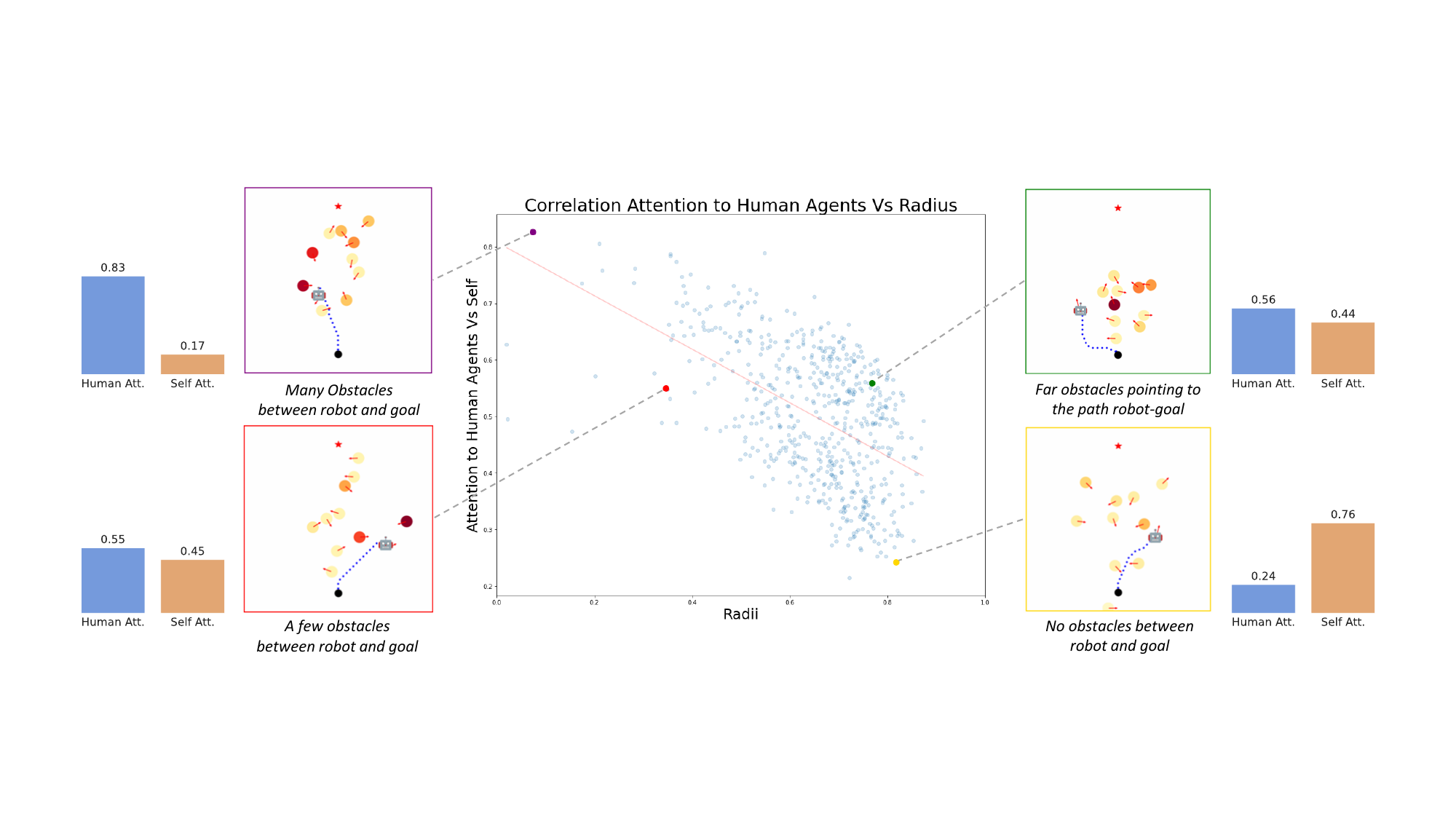}
    \caption{\textbf{Correlation analysis between hyperbolic radius and robot uncertainty.} Our hyperbolic embedding spaces inherently encode various navigation scenarios, ranging from simple to complex. Complex scenarios often require altering a planned trajectory or adjusting steering to avoid potential collisions. When dealing with many close obstacles and possible collisions, our policy obtains embeddings closer to the origin (\textit{purple} and \textit{red} frames in the Figure)
    . At the same time, easier and more certain cases (\textit{green} and \textit{yellow} frames) yield embeddings near the boundary of hyperbolic space. Hence, our approach comes with a simple way to measure how certain a robot is at any given time. 
    }
\label{fig:correlation}
\vspace{-0.3cm}
\end{figure*}

%% file: Sections/8-Limitations.tex
\section{Limitations}\label{limitations}
In Sec.~\ref{exp_results}, we show that our proposed model reports the best success rate at a fraction of the competitors' parameter count, which results in a decreased memory footprint (0.21 Vs. 1.38 MB for \modelname and SafeCrowdNav, respectively). However, we acknowledge that ours is not the faster model in terms of runtime (30.6 Vs. 13.4 msec for \modelname and SafeCrowdNav, respectively). This is due to the intrinsic complexities associated with the hyperbolic space computations. The primary hyperbolic operations (cf. Eqs.~\ref{expmapO},\ref{logmapO},\ref{PoincareDist}) are mathematically complex and less optimized in the current deep learning framework, which primarily caters to operations in Euclidean spaces.\\
Another limitation stems from employing simulations. 
\modelname has been tested in challenging realistic crowded simulations but the human agents follow a given ORCA~\cite{van2008orca} policy. Also, the robot does not elicit human reactions by design. Both aspects may be resolved by future more realistic simulations or by real-world deployment tests, having ensured the safety of the human participants.

%% file: Sections/7-Conclusions.tex
\section{Conclusions}

In this paper, we have introduced \modelname, a novel model for crowd navigation that exploits hyperbolic latent spaces to encode 
the environment into states which are natively hierarchical. 
Adhering to the hyperbolic learning framework allows \modelname to achieve state-of-the-art success rates and average return with a significant reduction in parameter count with respect to competitive baselines, ensuring that \modelname consistently devises safe paths in an efficient manner. We have shown that the hyperbolic framework comes with the additional benefit of greater interpretability, as monitoring the hyperbolic radius throughout a crowd navigation episode reveals insights about the complexity of the current state of the environment from the robot's perspective. \new{As future works, we plan to explore the generalization of our hyperbolic RL approach to a broader range of environments beyond just navigation tasks, as well as investigate how the hierarchical inductive bias of the hyperbolic space impacts learning in domains with more explicit multi-level structure.} 

\section*{\new{Acknowledgements}}

\new{We acknowledge financial support from the PNRR MUR project PE0000013-FAIR and from the Sapienza grant RG123188B3EF6A80 (CENTS).}